\pdfoutput=1
\documentclass[11pt]{article}

\usepackage[final]{acl}

\usepackage{times}
\usepackage{latexsym}

\usepackage[T1]{fontenc}
\usepackage[utf8]{inputenc}

\usepackage{microtype}

\usepackage{inconsolata}

\usepackage{graphicx}
\usepackage{caption}
\usepackage{booktabs}
\usepackage{graphicx}
\usepackage{multirow}
\usepackage{arydshln}
\usepackage{latexsym}
\usepackage{tcolorbox}
\usepackage{microtype}
\usepackage{subcaption}
\usepackage{inconsolata}
\usepackage{amsmath, amsfonts}
\usepackage{float}
\usepackage{dsfont}
\usepackage[table]{xcolor}
\usepackage{mathrsfs}

\newtheorem{definition}{Definition}

\title{Text Anomaly Detection with Simplified Isolation Kernel}

\author{
 \textbf{Yang Cao\textsuperscript{1,2,3,4,5}},
 \textbf{Sikun Yang\textsuperscript{1,2,3,4}\thanks{Corresponding author.}},
 \textbf{Yujiu Yang\textsuperscript{5}},
 \textbf{Lianyong Qi\textsuperscript{6}},
 \textbf{Ming Liu\textsuperscript{7}},
\\
 \textsuperscript{1}School of Computing and Information Technology, Great Bay University, China \\
 \textsuperscript{2}Great Bay Institute for Advanced Study, Great Bay University, China \\
  \textsuperscript{3}Guangdong Provincial Key Laboratory of Mathematical and Neural Dynamical Systems\\
 \textsuperscript{4}Dongguan Key Laboratory for Intelligence and Information Technology, China \\
 \textsuperscript{5}Tsinghua Shenzhen International Graduate School, Tsinghua University, China \\
 \textsuperscript{6}China University of Petroleum (East China), China \\
 \textsuperscript{7}School of IT, Deakin University, Australia
\\
 \small{
   \textbf{Correspondence:} \href{charles.cao@ieee.org}{charles.cao@ieee.org}
 }
}

\begin{document}
\maketitle
\begin{abstract}

Two-step approaches combining pre-trained large language model embeddings and anomaly detectors demonstrate strong performance in text anomaly detection by leveraging rich semantic representations. However, high-dimensional dense embeddings extracted by large language models pose challenges due to substantial memory requirements and high computation time. To address this challenge, we introduce the Simplified Isolation Kernel (SIK), which maps high-dimensional dense embeddings to lower-dimensional sparse representations while preserving crucial anomaly characteristics. SIK has linear time complexity and significantly reduces space complexity through its innovative boundary-focused feature mapping. Experiments across 7 datasets demonstrate that SIK achieves better detection performance than 11 state-of-the-art (SOTA) anomaly detection algorithms while maintaining computational efficiency and low memory cost. All code and demonstrations are available at \url{https://github.com/charles-cao/SIK}.

\end{abstract}

\section{Introduction}

Text anomaly detection (TAD) plays a crucial role in many applications, including content moderation, fraud detection, and cybersecurity threat analysis~\cite{pang2021deep}. TAD involves identifying textual instances that significantly deviate from the norm, which could indicate potential security threats, novel information, or content requiring special attention~\cite{cao2025tad}. With the exponential growth of digital text data, developing effective and efficient text anomaly detection methods has become increasingly important.

Text anomaly detection methods generally fall into two categories: end-to-end approaches and two-step approaches~\cite{li2024nlp}. End-to-end approaches integrate representation learning and anomaly detection into unified frameworks. 
However, they require substantial data for each specific domain and demand complete retraining when deployed to new domains. Their poor generalization across different text corpora makes them impractical for many real-world scenarios where anomalies vary across contexts and domains~\cite{malik2024deep}.

Recent advances in large language models have created powerful embedding techniques that extract meaningful feature representations from various data types. These modular approaches to anomaly detection follow a two-step process~\cite{li2024nlp}: 1) extracting dense vector embeddings that capture semantic relationships and contextual information from the raw data; 2) applying traditional anomaly detection algorithms on these embeddings. This approach leverages pre-trained models to directly extract features, eliminating the need for model retraining and significantly improving computational efficiency.

Isolation-based anomaly detection methods have demonstrated exceptional performance in text anomaly detection tasks \cite{cao2025tad}. The latest method Isolation Kernel (IK)~\cite{ting2020isolation} has also been widely applied to anomaly detection in time series~\cite{cao2024detecting}, streaming data~\cite{cao2025revisiting}, and graph domains~\cite{zhuang2023subgraph} due to its data-dependent characteristics~\cite{cao2025anomaly}. However, IK requires mapping data to a high-dimensional space before performing anomaly detection, which significantly increases computational time and memory cost. The limitations become particularly problematic in the context of modern large language models (LLMs), where the embeddings extracted by LLMs already possess inherently high dimensionality, often reaching several hundred or thousand dimensions~\cite{devlin2019bert}.

% Modern text embeddings from advanced language models are increasingly high-dimensional, often reaching several hundred or thousand dimensions~\cite{devlin2019bert}. While these high-dimensional representations capture richer semantic information, computational time and memory requirements also scale with dimensionality, creating efficiency barriers for large-scale text corpora processing.

To address these limitations, we propose the Simplified Isolation Kernel (SIK). SIK effectively maps high-dimensional dense embeddings to a lower-dimensional sparse representation while preserving crucial information for anomaly detection. 
The core intuition of SIK is that for text anomaly detection tasks, we only need to focus on the dissimilarity between normal and anomalous samples, while the dissimilarity among normal samples can be ignored.
% Both SIK and IK employ the same hyperspheres space partitioning mechanism. However, SIK only needs to determine whether a point falls into any hypersphere, without needing to identify which specific hypersphere it falls into.
The key contributions of our work are:
\begin{itemize}
    \item We propose Simplified Isolation Kernel (SIK), which maps high-dimensional embeddings to sparse representations by focusing only on whether a point lies outside the normal data boundary and discarding redundant similarity information among normal instances.
    \item SIK achieves linear time and space complexity, enabling scalable and efficient anomaly detection while maintaining high performance.
    \item Empirical evaluation demonstrates that SIK has better detection performance than existing methods across multiple domains and embeddings.
\end{itemize}

% The rest of the paper is organized as follows. Section~\ref{rw} describes the related work in TAD tasks. Section~\ref{pre} provides the definitions of problem, isolation kernel and isolation distribution kernel. Section~\ref{method} introduces the proposed method SIK. Section~\ref{exp} gives the experimental results and conclusions are provided in the last section.

\section{Related Work}\label{rw}

\subsection{Text Representations}

The evolution of text representation techniques has been pivotal in advancing natural language processing. Early approaches such as TF-IDF~\cite{salton1988term} created sparse vector representations that, while computationally efficient, were limited in capturing semantic relationships between words. This limitation was partially addressed by Word2Vec~\cite{mikolov2013efficient} and GloVe~\cite{pennington2014glove}, which generated dense continuous vector spaces based on word co-occurrence patterns, though they still assigned static representations regardless of contextual usage. The field subsequently progressed toward contextualized embeddings with ELMo~\cite{peters-etal-2018-deep} and transformer-based architectures like BERT~\cite{devlin2019bert}, which revolutionized NLP by using bidirectional attention mechanisms to produce context-sensitive representations.

The landscape of text representation was further transformed by the emergence of large language models (LLMs) exemplified by GPT~\cite{brown2020language}. LLMs are trained on vast and diverse corpora, generating remarkably expressive embeddings that capture deep semantic relationships and generative text properties. 

\subsection{End-to-end TAD Approaches}

End-to-end approaches integrate representation learning and anomaly detection into unified frameworks. Early neural methods primarily relied on autoencoder architectures to model normal text patterns, identifying anomalies through reconstruction errors~\cite{manevitz2007one}.

More recent innovations have shifted toward transformer-based architectures for text anomaly detection. CVDD~\cite{ruff2019self} detects anomalies by learning multiple context vectors through self-attention mechanisms on word embeddings, then identifying outliers based on the distance between text representations and these context vectors. DATE~\cite{manolache2021date} identifies replaced tokens and recognizes which masking pattern was applied to normal text, then scoring anomalies based on the model's uncertainty when processing unfamiliar patterns. FATE~\cite{das2023few} leverages a small number of labeled anomalous examples along with a deviation learning approach, where normal texts are pushed to match reference scores from a prior distribution while anomalous texts are forced to deviate significantly.

\subsection{Two-step Approaches}

Based on the text embeddings, traditional anomaly detection methods can be applied and they are classified into several distinct approaches, each with specific strengths for different data distributions and anomaly types. 

Density-based methods like LOF~\cite{breunig2000lof} identify outliers by measuring local density deviations relative to neighboring points. Isolation techniques, including iForest~\cite{liu2008isolation, liu2012isolation} and iNNE~\cite{bandaragoda2018isolation}, operate on the principle that anomalies are sparse and distinctive, using space partitioning strategies where anomalous points require fewer partitions or are assigned to larger or out of hyperspheres.

Statistical approaches detect anomalies through their deviation from established data distributions, with ECOD~\cite{li2022ecod} utilizing cumulative distribution functions for efficient scoring and COPOD~\cite{li2020copod} employing copulas to effectively model dependencies in multivariate scenarios. Meanwhile, deep learning methods have emerged as powerful tools for capturing complex, nonlinear patterns in data. Models such as Deep SVDD~\cite{ruff2018deep} and LUNAR~\cite{goodge2022lunar} learn representations from normal instances and identify anomalies as significant deviations from these learned patterns, though they typically demand substantial training data and computational resources to achieve optimal performance.

\section{Preliminaries}\label{pre}

Table~\ref{tab:notation} presents the key symbols and notations used in this paper.

\begin{table}[!htbp]
    \centering
    \caption{Key symbols and notations}
    \resizebox{0.48\textwidth}{!}{
    \begin{tabular}{ll}
    \toprule
    % $x$ or $y$   & An embedding point extracted by language models \\
    % $P_Y$     & Probability distribution that generates the set $Y \subseteq \mathbb{R}^d$\\
    $\kappa_{I}$ & Isolation Kernel \\
    $\widehat{\mathcal{K}}_I$ & Isolation Distributional Kernel \\
    $\Phi$    & Feature map of Isolation Kernel\\
    $\widehat{\Phi}$ & Kernel mean map of Isolation Distributional Kernel \\
    $S_{IK}$ & Anomaly scores of Isolation Kernel \\
    $\kappa_{S}$ & Simplified Isolation Kernel \\
    $\phi$    & Feature map of Simplified Isolation Kernel \\
    $S_{SIK}$ & Anomaly scores of Simplified Isolation Kernel \\
    \bottomrule
    \end{tabular}}
    \label{tab:notation}
\end{table}

% \textcolor{red}{Need a table of symbols that describe each key symbols used. Many symbols are not defined, e.g., $P_X$}

\subsection{Problem Definition}
% \textcolor{red}{Need a definition since the title says so.}

Text anomalies are instances that significantly deviate from established patterns within a document collection. These anomalies may manifest as unusual topics, atypical linguistic structures, domain-specific terminology, or deliberately manipulated content such as spam, misinformation, or hate speech. Detecting such anomalies serves valuable purposes in content moderation, deception detection, and security surveillance.

Let $D = \{x_1, x_2, \dots, x_N\}$ represent a collection of $N$ text documents, where each document $x_i$ is a sequence of lexical elements: $x_i = \{token_1, token_2, \dots, token_{L_i}\}$,
with $L_i$ denoting the document's length in tokens.

The core objective in text anomaly detection is distinguishing $D$ into two disjoint subsets: $D_{\text{normal}}$ and $D_{\text{anomalous}}$, where $D_{\text{anomalous}}$ contains documents that substantially differ from the dominant patterns exhibited by $D_{\text{normal}} = D \setminus D_{\text{anomalous}}$.

% The primary challenge lies in developing an effective function $g$ that can accurately distinguish between normal and anomalous documents without labels, while remaining robust across diverse textual distributions and high-dimensional embedding spaces.

\subsection{Isolation Kernel (IK)}

Isolation Kernel (IK)~\cite{ting2018isolation} is a data-dependent kernel that derives directly from the dataset without a learning process. It has been used in many different anomaly detection application scenarios, including time series~\cite{ting2022new, ting2024new}, streaming data~\cite{cao2025revisiting} and graphs~\cite{zhuang2023subgraph}, etc. The fundamental principle behind IK involves estimating the probability that two points will be assigned to the same partition through a data space partitioning strategy. Previous implementations of IK have utilized various partitioning mechanisms, including iForest~\cite{ting2018isolation}, hypersphere~\cite{ting2020isolation}, and Voronoi diagram~\cite{qin2019nearest} approaches. In this paper, we specifically employ the hypersphere partitioning strategy, with detailed methodological explanations provided in Section~\ref{sp}. 

Let $D \subset \mathcal{X} \subseteq \mathbb{R}^d$ be a dataset sampled from an unknown distribution $P_D$, and $\mathds{H}_\psi(D)$ denote the set of all partitionings $H$ that are admissible from $\mathcal{D} \subset D$, where each sample point $z \in \mathcal{D}$ has an equal probability of being selected from $\mathcal{D}$, and $|\mathcal{D}| = \psi$.

The key idea of IK is to use $\psi$ random sample points $z$ to partition the data space, and the detailed partitioning strategy is provided in Section~\ref{sp}. The similarity between two points $x$ and $y$ is the is the number of times that both of them fall into the same partition $\theta[z]$ across $t$ partitionings.   

\begin{definition}\label{IKernel} 
\cite{ting2018isolation, qin2019nearest} For any two points $x,y \in \mathbb{R}^d$,
	Isolation Kernel of $x$ and $y$ is defined to be
	the expectation taken over the probability distribution on all partitionings $H \in \mathds{H}_\psi(D)$ that both $x$ and $y$  
    fall into the same isolating partition $\theta[z] \in H$, 
    $z \in \mathcal{D} \subset D$:
	\begin{eqnarray}
        \kappa_I(x,y\ |\ D)   &=&  {\mathbb E}_{\mathds{H}_\psi(D)} [\mathds{1}(x,y \in \theta[z]\ ] \nonumber \\
         &=& \frac{1}{t}\langle\Phi(x), \Phi(y)\rangle,
	\end{eqnarray}
    where $\mathds{1}(\cdot)$ be an indicator function. 
\end{definition}

Each Partitioning $H_i (i = 1,\dots,t)$ is a complete division of the feature space using $\psi$ randomly sampled hyperspheres, and creates $\psi+1$ regions: $\psi$ hypersphere interiors and the exterior region. Each partition $\theta_j (j=1,\dots,\psi)$ is an individual hypersphere region within partitioning $H_i$. The union of these $\psi$ hyperspheres creates a boundary that divides the data space into two parts: inside the boundary (covered by at least one hypersphere) and outside the boundary (not covered by any hypersphere).

\begin{definition}[Feature Map of IK]~\cite{ting2020isolation}
    IK maps each point $x$ to a $t \times \psi$ dimensions binary feature map $\Phi: x \mapsto \{0,1\}^{t \times \psi}$. Specifically, given $t$ partitionings, for each partitioning $H_i$ ($i = 1, \ldots, t$), IK creates a $\psi$-dimensional binary column vector $\Phi_{i}(x)$ where each dimension corresponds to one of the $\psi$ partitions in $H_i$. The $j$-th component of this vector is:
\begin{equation} 
\Phi_{i,j}(x) = \mathds{1}(x \in \theta_j \mid \theta_j \in H_i),
\end{equation}
where $j = 1, \ldots, \psi$. This indicates whether point $x$ falls inside partition $\theta_j$ in partitioning $H_i$. The final representation $\Phi(x)$ is the concatenation of all vectors: $\Phi_{1}(x), \ldots, \Phi_{t}(x)$. 
\end{definition}

% \textcolor{red}{Get rid of subscript IK in $\Phi$.}

\subsection{Isolation Distributioal Kernel (IDK)}
Based on the same framework of Kernel Mean Embedding (KME)~\cite{muandet2017kernel}, IK has been used as the foundation to develop a distributional kernel called Isolation Distributional Kernel (IDK)~\cite{ting2020isolation}. IDK specifically measures the similarity between two distributions rather than just between individual points. For text anomaly detection, the anomaly score of each point can be computed by measuring the similarity between each point and the whole data distribution.

\begin{definition}
Isolation Distributional Kernel of a point distribution $\mathcal{P}_{x}$ and a distribution $\mathcal{P}_Y$ is: 
\begin{equation}
{\widehat{\mathcal{K}}_I}(\mathcal{P}_{x},\mathcal{P}_Y\ |\ D) 
  =  \frac{1}{t} \left< \Phi(\mathcal{P}_{x}|D), \widehat{\Phi}(\mathcal{P}_Y|D) \right> , 
\end{equation}
where  $\widehat{\Phi}(\mathcal{P}_Y|D)  =  \frac{1}{|Y|} \sum_{y \in Y} \Phi(y|D)$ is the kernel mean map. 
 \end{definition}

\section{Methodology}\label{method}

\begin{figure*}[!htbp]
    \centering
    \includegraphics[width=0.75\linewidth]{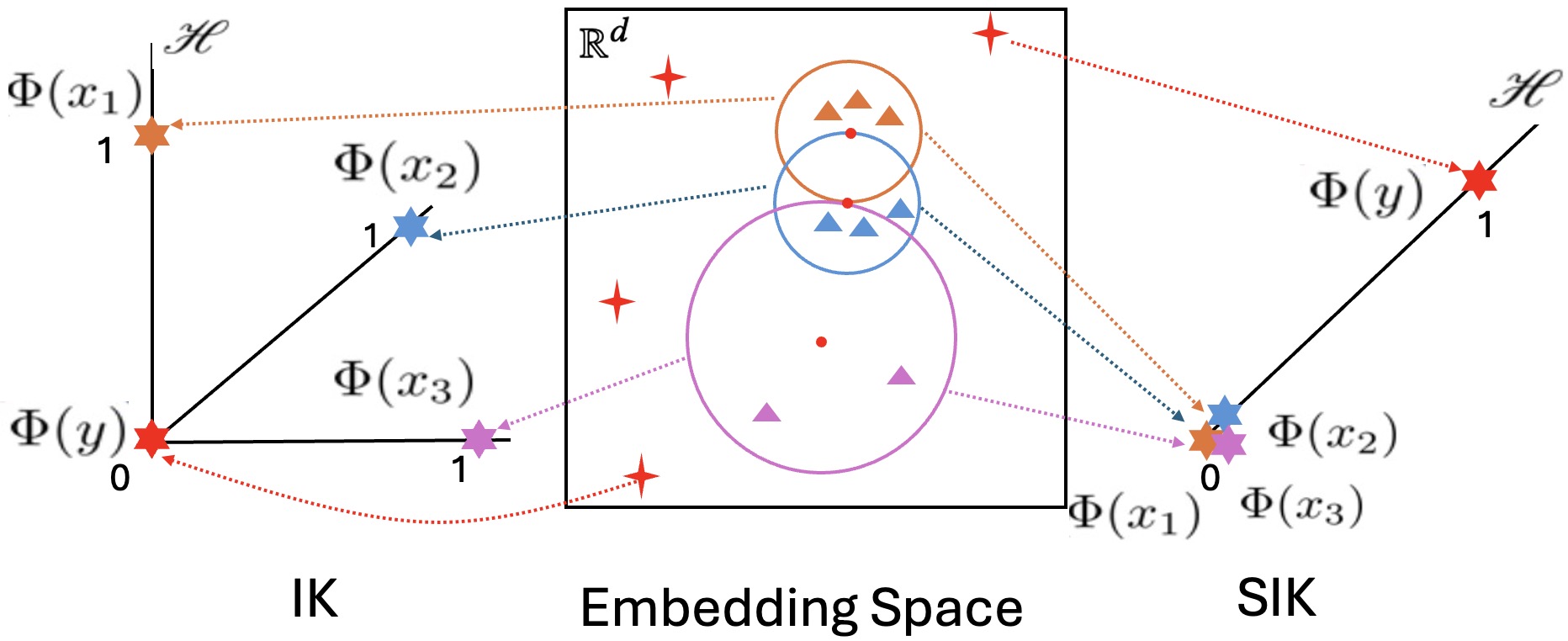}
    \caption{An illustration of feature maps of IK and SIK with one partitioning ($t=1$) of 3 hyperspheres. Each center of a hypersphere is at a point $z \in D$ where $\psi = 3$ are randomly selected from the given dataset $D$. When a point $x$ falls into an overlapping region, $x$ is regarded as being in the hypersphere whose center is closer to $x$. On the left IK feature space, $x$ has a 3-dimensional feature vector. On the right SIK feature space, $x$  has only a 1-dimensional feature vector.}
    % \textcolor{red}{Have a clear space between IK feature space and the middle figure. IBK $=>$ SIK}}
    \label{fig:demo}
\end{figure*}

For text anomaly detection, we follow a two-step approach: text documents are first transformed into dense vector embeddings that capture semantic relationships. These embeddings can be generated using pre-trained language models, which encode contextual information and linguistic patterns. The quality of embeddings significantly influences downstream anomaly detection performance. Once documents are embedded in a high-dimensional space, traditional anomaly detection algorithms can be applied to identify outliers. However, these embeddings typically exist in high-dimensional spaces and applying anomaly detection algorithms to such high-dimensional data creates substantial computational challenges. 
We introduce Simplified Isolation Kernel (SIK) to address these challenges, the key steps are shown in the following subsections.
% by mapping high-dimensional dense embeddings to low-dimensional sparse representations while preserving essential information for anomaly detection. 

\subsection{Space Partitioning}\label{sp}

The proposed SIK employs a hypersphere-based space partitioning mechanism, following the same approach as used in iNNE~\cite{bandaragoda2018isolation} and IDK~\cite{ting2020isolation}. The fundamental idea is to create a collection of hyperspheres that adapt to the local density of the data, which enables effective anomaly detection across regions of various densities.

\begin{definition}[Hypersphere Partionings]
    Each point $z \in \mathcal{D}$ is isolated from the rest of the points in $\mathcal{D}$ by building hyperspheres $\theta[z] \in H$ centered at $z$. The radius of this hypersphere is determined by the distance between $z$ and its nearest neighbor in $\mathcal{D} \setminus \{z\}$. Each partitioning $H$ consists of $\psi$ hyperspheres and the region that is not covered by these hyperspheres. For stability, $t$ different partitionings $H_i$, $i = 1\dots t$ are generated, each based on a different random subset $\mathcal{D}_i \subset D$. 
\end{definition}

This partitioning mechanism naturally adapts to the underlying data distribution. In dense regions, the resulting hyperspheres have short radii, since nearest neighbors are typically close. Conversely, in sparse regions, the hyperspheres have long radii because nearest neighbors are farther apart. This data-dependent property is crucial for effective anomaly detection, as it creates adaptive partitionings that can appropriately in both dense and sparse regions.

Unlike fixed-radii approaches that may struggle with varying data densities, this adaptive mechanism provides appropriate coverage across the entire feature space. It avoids overfitting in dense regions (where a fixed small radius would create too many partitions) and underfitting in sparse regions (where a fixed large radius might miss important structural details). 

The first difference between IK and SIK lies in their fundamental approaches, despite using the same hypersphere partitioning mechanism. Both methods agree that points falling outside hyperspheres are more likely to be abnormal. However, IK focuses on pairwise similarity measurement, calculating how many times two points fall into the same specific hypersphere to determine their similarity. In contrast, SIK adopts a boundary-based perspective, constructs a decision boundary using multiple hyperspheres. In SIK, we assume that the higher the frequency of a point falling outside these boundaries, the more likely it is to be abnormal. SIK deliberately ignores the specific position of points within the boundaries, as this information is less relevant for anomaly detection purposes.

\subsection{Feature Map}\label{fm}

The second key difference between IK and our proposed SIK lies in their feature representation approaches. 

While IK achieves linear time complexity, it creates high-dimensional feature representations by tracking exactly which hypersphere contains each point. This approach becomes problematic for large-scale text anomaly detection, where input data (e.g., 768-dimensional BERT embeddings) is already high-dimensional. IK feature mapping further expands this dimensionality, creating prohibitive memory requirements. Figure~\ref{fig:demo} left shows that IK maps all points to 3-dimensions ($\psi=3$) space in one partitioning.

SIK addresses this limitation through a more compact feature representation, and the key insight is that anomaly detection doesn't require determining that a point falls into which specific hypersphere, but is sufficient to know whether a point falls into the boundary. This simplification significantly reduces the feature dimension while preserving the critical information needed for anomaly detection. 
% Figure~\ref{fig:demo} right shows that all points are mapped to a 1-dimensional space, points in the boundary are mapped to the original of RKHS, and others are mapped to 1.

\begin{definition}[Feature Map of SIK]
Given a point $x\in \mathbb{R}^d$, the feature map $\phi: x \mapsto \{0,1\}^t$ of SIK is a $t$-dimensional binary column vector, where each component $H_i$ ($i = 1, \ldots, t$) indicates whether $x$ falls outside all hyperspheres $\theta\in H_i$:
\begin{equation}
\phi_{i}(x) = \mathds{1}(x \notin \theta| \theta \in H_i). 
\end{equation}
% where $\mathds{1}(\cdot)$ is the indicator function.
\end{definition}

% \textcolor{red}{Get rid of subscript SIK; and use a different symbol (e.g., $\phi$) instead of $\Phi$.}

The SIK kernel function between two points $x$ and $y$ can be formally defined as:

\begin{align}
\kappa_{S}(x, y) &= {\mathbb E}_{\mathds{H}_\psi(D)} [\mathds{1}(x,y \notin \theta\ | \ \theta \in H_i)] \nonumber \\
&=\frac{1}{t}\sum_{i=1}^{t} [\mathds{1}(x,y \notin \theta\ | \ \theta \in H_i)] \nonumber \\
% &=\frac{1}{t}\sum_{i=1}^{t}\sum_{\theta \in H_i}^{}\mathds{1}(x \notin \theta)\mathds{1}(y \notin \theta) \nonumber\\
&= \frac{1}{t}\langle\phi(x), \phi(y)\rangle.
% &= \sum_{i=1}^{t} \Phi_{SIK}(x) \cdot \Phi_{SIK}(y)
\label{sik}
\end{align}

Figure~\ref{fig:demo} illustrates how both methods map data points differently in the Reproducing Kernel Hilbert Space (RKHS) $\mathscr{H}$. IK tracks the exact hypersphere membership for each partitioning, creating high dimensional feature vectors that specify which particular hypersphere contains each point. For example, with $t=2$ partitionings and $\psi=3$ hyperspheres per partitioning, IK produces a 6-dimensional feature vector like [0,1,0,0,0,0], indicating the point falls in the second hypersphere of the first partitioning and outside all hyperspheres in the second partitioning. In contrast, SIK adopts a boundary-focused approach that only distinguishes whether a point falls inside any hypersphere versus outside all hyperspheres within each partitioning. Using the same example, SIK generates a compact 2-dimensional feature vector [0,1], indicating the point is inside one hypersphere in the first partitioning but outside all hyperspheres in the second partitioning. This simplification reduces feature dimensionality from $\psi t$ to $t$ while preserving the essential boundary information needed for anomaly detection, as anomalous points are primarily characterized by their tendency to fall outside normal data boundaries rather than their specific hypersphere assignments.

\subsection{Anomaly Scores Calculation}
Unlike traditional kernel functions that typically measure similarity between points, SIK quantifies how many times two points simultaneously fall outside all hyperspheres across multiple partitionings. When both points consistently fall outside all hyperspheres, they have a high SIK value. This characteristic allows us to compute anomaly scores by measuring the similarity between each point's feature vector and a reference anomaly vector. And the ideal anomaly point $\mathcal{A}$ should fall outside all hyperspheres in all partitionings, where its vector will be $[1,\dots,1]$. Thus, the anomaly score of each point $x$ can be defined as the similarity between the point $x$ and the ideal reference anomaly point $\mathcal{A}$. 
% \textcolor{red}{Need to make it clearer. ideal anomaly point: $\mathcal{A}$ or $Y$?}

\begin{definition}[Anomaly Score]
Given the binary feature representation $\phi(x) \in \{0, 1\}^t$ for a point $x \in \mathbb{R}^d$, and let $\mathcal{A}$ be an reference anomaly point with $\phi(\mathcal{A}) = [1,\dots,1]$, the anomaly score is defined as:

\begin{equation}
S_{SIK}(x) = \frac{1}{t} \langle \phi(x), \phi(\mathcal{A}) \rangle,  
\end{equation}

\noindent the range of $S_{SIK}$ is $[0,1]$ since $0 \leq S_{SIK}(x) \leq t$.

\end{definition}

Under this formulation, points with scores approaching 1 after normalization) are more likely to be anomalies as they have higher similarity to the ideal anomaly, while normal points typically have scores closer to 0.

Since SIK feature map consists of 0 and 1, the anomaly score of point $x$ are equivalent to the Hamming distance between $\phi(x)$ and the origin $[0,\ldots,0]$, which 
quantifies the degree that a point is isolated from regions of normal data. In addition, the anomaly scores can be equivalently expressed in terms of $L_0$ and $L_1$ norm as well.

The score calculation method can also be applied to the IK feature map:

\begin{equation}
S_{IK}(x) = 1 - \frac{1}{t} \parallel \Phi(x) \parallel, 
\end{equation}
where $\parallel \cdot \parallel$ can be either L0 or L1 norm.

It is worth noting that $S_{IK}(x)$ is equal to $S_{SIK}(x)$ since both methods essentially count how many times a point falls outside all hyperspheres across the $t$ partitionings, directly measuring its degree of isolation from normal data regions.

\subsection{Is SIK a Valid Kernel?}

According to Mercer's theorem, a symmetric function $\kappa: \mathcal{X} \times \mathcal{X} \rightarrow \mathbb{R}$ is a valid kernel only if it is positive semi-definite and symmetric~\cite {christmann2008support}. We demonstrate that SIK satisfies both requirements.

% The SIK kernel is defined as:
% \begin{equation}
% \kappa_{\text{SIK}}(x, y) = \langle \Phi_{\text{SIK}}(x), \Phi_{\text{SIK}}(y) \rangle
% \end{equation}

Based on Equation~\ref{sik}, for symmetry, we observe that:
\begin{align}
\kappa_S(x, y) &= \frac{1}{t}\langle \phi(x), \phi(y) \rangle \nonumber \\
&= \frac{1}{t}\langle \phi(y), \phi(x) \rangle \nonumber \\
&= \kappa_S(y, x)
\end{align}
This confirms SIK satisfies symmetry.

For positive semi-definiteness, Mercer's theorem requires that for any data points $x_1,...,x_n \in \mathbb{R}^d$ and any real coefficients $\alpha_1,...,\alpha_n \in \mathbb{R}$:
\begin{equation}
\sum_{i=1}^{n} \sum_{j=1}^{n} \alpha_i \alpha_j \kappa_{S}(x_i, x_j) \geq 0
\label{psd}
\end{equation}

By the properties of inner products and the fact that $\kappa_S(x_i, x_j) = \langle \phi(x_i), \phi(x_j) \rangle$, rewritten Equation~\ref{psd} as:
\begin{equation}
\left\langle \sum_{i=1}^{n} \alpha_i \phi_S(x_i), \sum_{j=1}^{n} \alpha_j \phi(x_j) \right\rangle \geq 0
\end{equation}

Since both summations represent the same vector in feature space, this simplifies to:
\begin{equation}
\left\| \sum_{i=1}^{n} \alpha_i \phi(x_i) \right\|^2 \geq 0
\end{equation}

This inequality always holds since a squared norm is non negative. It is important to note that the feature map $\phi$ maps input points to binary vectors, which further supports the positive semi-definiteness property. Therefore, by Mercer's theorem, SIK is a valid kernel function.

\subsection{Relationship to SiNNE}

The closest work to SIK is SiNNE~\cite{samariya2020new}, which also employs hypersphere-based partitioning and binary feature representation where points inside hyperspheres are assigned 0 and points outside are assigned 1. SiNNE was developed as a simplified version of iNNE to address computational efficiency issues in outlying aspect mining—a task that aims to identify which specific feature subsets (aspects) make a given data point anomalous, essentially explaining why a point is an outlier rather than detecting whether it is an outlier. SiNNE operates based on the intuitive assumption that anomalous points should fall outside hyperspheres more frequently than normal points, but provides no theoretical justification for why this assumption holds or guarantees its effectiveness. In contrast, our Simplified Isolation Kernel emerges from rigorous kernel theory, where we prove that SIK satisfies both symmetry and positive semi-definiteness properties, establishing it as a valid kernel function. This kernel foundation provides theoretical guarantees for anomaly detection effectiveness, as the kernel essentially measures similarity between data points—anomalous points naturally exhibit low similarity to the majority of normal points in the kernel space. 
% Furthermore, SIK addresses a fundamentally different problem: detecting anomalous text documents rather than explaining which features make them anomalous, with design specifically targeting the computational challenges of applying Isolation Kernel to high-dimensional dense text embeddings from large language models.

\section{Experiments}\label{exp}

\subsection{Experimental Setups}

The experiments are conducted using the same benchmark datasets and embeddings from the benchmark NLP-ADBench~\cite{li2024nlp}. The embeddings are extracted via BERT~\cite{devlin2019bert} and OpenAI (text-embedding-3-large)~\cite{openai2023new_embeddings} models as specified in NLP-ADBench. Statistical information of the experiment datasets is summarized in Table~\ref{tab:datasets}.
AUROC (Area Under the Receiver Operating Characteristic Curve) is adopted as the evaluation metric. Each experiment is repeated 5 times with average results reported to mitigate randomness. 

We utilized 8 traditional methods (LOF, iForest, ECOD, DeepSVDD, Autoencoder, LUNAR, INNE and IDK) sourced from the \textit{PyOD} library \cite{zhao2019pyod} on the extracted embeddings. The hyperparameters of nearest neighbors for LOF and LUNAR are searched in $\{5, 10, 20, 40\}$. For iForest, iNNE, IDK and SIK, $\psi$ is searched in $\{32, 64, 128, 256, 512\}$ and with defalt $t=200$. For Autoencoder and DeepSVDD, the hyperparameters of hidden neurons are searched in $\{[128, 64], [64, 32], [32, 16]\}$.

For comparative analysis, 3 end-to-end methods (CVDD, DATE and FATE) are included in this paper, and their performance is directly referenced from the NLP-ADBench~\cite{li2024nlp} due to the same datasets.

\begin{table}[!htbp]
\caption{Statistical information of datasets}
\label{tab:datasets}
\resizebox{0.48\textwidth}{!}{%
\begin{tabular}{l|cccll}
\hline
Dataset     & \# Samples & \# Ano. & \% Ano. & Train  & Test   \\ \hline
EmailSpam   & 3578       & 146     & 4.08    & 2402   & 1176   \\
SMSSpam     & 4969       & 144     & 2.89    & 3162   & 1510   \\
BBCNews     & 1785       & 62      & 3.47    & 1206   & 579    \\
AGNews      & 98207      & 3780    & 3.85    & 66098  & 32109  \\
N24News     & 59822      & 1828    & 3.06    & 40595  & 19227  \\
MovieReview & 26369      & 1487    & 5.64    & 17417  & 8952   \\
YelpReview  & 316924     & 17938   & 5.66    & 209290 & 107634 \\ \hline
\end{tabular}%
}
\end{table}

\subsection{Empirical Evaluation}

\begin{table*}[!htbp]
\caption{Evaluation results across 7 datasets in terms of AUROC. SIK results are shown with a shadow background, and the best result on each dataset is in bold. }
% \textcolor{red}{IK or IDK?}}
\label{tab:result}
\resizebox{\textwidth}{!}{%
\begin{tabular}{llllllll}
\hline
\textbf{Algorithms} & \textbf{Email\_Spam} & \textbf{SMS\_Spam} & \textbf{BBC\_News} & \textbf{AG\_News} & \textbf{N24News} & \textbf{Movie}  & \textbf{Yelp}   \\ \hline
CVDD                & 0.9340               & 0.4782             & 0.7221             & 0.6046            & 0.7507           & 0.4895          & 0.5345          \\
DATE                & 0.9697               & \textbf{0.9398}    & 0.9030             & 0.8120            & 0.7493           & 0.5185          & 0.6092          \\
FATE                & 0.9061               & 0.6262             & 0.9310             & 0.7756            & 0.8073           & 0.5289          & 0.5945          \\ \hline
BERT+LOF            & 0.7793               & 0.7642             & 0.9412             & 0.7643            & 0.6991           & 0.5253          & 0.6842          \\
BERT+iForest        & 0.7599               & 0.6544             & 0.7394             & 0.6760            & 0.5804           & 0.4624          & 0.6222          \\
BERT+ECOD           & 0.7427               & 0.6164             & 0.7302             & 0.6578            & 0.5363           & 0.4434          & 0.6204          \\
BERT+DeepSVDD       & 0.6200               & 0.5765             & 0.6841             & 0.6290            & 0.5373           & 0.4732          & 0.6066          \\
BERT+AE             & 0.8067               & 0.7526             & 0.9117             & 0.7491            & 0.6465           & 0.4975          & 0.6728          \\
BERT+LUNAR          & 0.8340               & 0.7474             & 0.9404             & 0.7832            & 0.6589           & 0.4806          & 0.6721          \\
BERT+INNE           & 0.8531               & 0.7528             & 0.9235             & 0.7761            & 0.6360           & 0.5125          & 0.6861          \\
BERT+IDK             & 0.8649               & 0.7703             & 0.9473             & 0.7805            & 0.6625           & 0.5131          & 0.6829          \\
\cellcolor[gray]{0.9}BERT+SIK   & \cellcolor[gray]{0.9}0.8705               & \cellcolor[gray]{0.9}0.7719             & \cellcolor[gray]{0.9}0.9414             & \cellcolor[gray]{0.9}0.7755            & \cellcolor[gray]{0.9}0.6689           & \cellcolor[gray]{0.9}0.5264          & \cellcolor[gray]{0.9}0.6840          \\ \hline
OpenAI+LOF          & 0.9726               & 0.9032             & 0.9671             & 0.9013            & 0.8160           & 0.6731          & 0.7694          \\
OpenAI+iForest      & 0.5425               & 0.6131             & 0.6468             & 0.5364            & 0.5289           & 0.5886          & 0.5401          \\
OpenAI+ECOD         & 0.8926               & 0.6155             & 0.7780             & 0.7260            & 0.6179           & \textbf{0.6933} & 0.7706          \\
OpenAI+DeepSVDD     & 0.5291               & 0.5238             & 0.6010             & 0.5272            & 0.5885           & 0.5318          & 0.4808          \\
OpenAI+AE           & 0.6826               & 0.7933             & 0.9645             & 0.8684            & 0.7504           & 0.6597          & 0.7568          \\
OpenAI+LUNAR        & 0.9590               & 0.7855             & 0.9773             & \textbf{0.9309}   & 0.8324           & 0.6781          & \textbf{0.7984} \\
OpenAI+INNE         & 0.9727               & 0.8688             & 0.9833             & 0.8701            & 0.8067           & 0.6668          & 0.7367          \\
OpenAI+IDK           & 0.9531               & 0.8615             & 0.9797             & 0.8855            & 0.8290           & 0.6290          & 0.6688          \\
\cellcolor[gray]{0.9}OpenAI+SIK          & \cellcolor[gray]{0.9}\textbf{0.9729}      & \cellcolor[gray]{0.9}0.8967             & \cellcolor[gray]{0.9}\textbf{0.9844}    & \cellcolor[gray]{0.9}0.8904            & \cellcolor[gray]{0.9}\textbf{0.8343}  & \cellcolor[gray]{0.9}0.6634          & \cellcolor[gray]{0.9}0.7345          \\ \hline
\end{tabular}%
}
\end{table*}

Table~\ref{tab:result} presents the AUROC scores of all baseline methods across the 7 datasets. 

\begin{figure}[!htbp]
     \centering
     \begin{subfigure}[b]{0.48\textwidth}
         \centering
         \includegraphics[width=\textwidth]{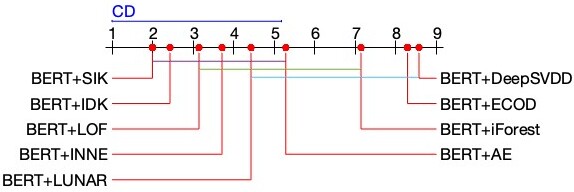}
         \caption{BERT-based}
         \label{bert}
     \end{subfigure}
     
     \begin{subfigure}[b]{0.48\textwidth}
         \centering
         \includegraphics[width=\textwidth]{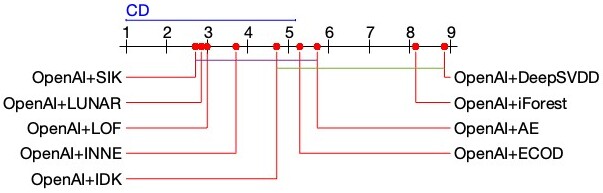}
         \caption{OpenAI-based}
         \label{openai}
     \end{subfigure}
     
     \begin{subfigure}[b]{0.4\textwidth}
         \centering
         \includegraphics[width=\textwidth]{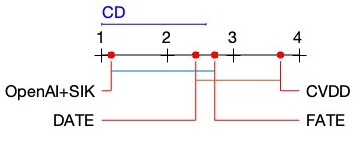}
         \caption{End-to-End}
         \label{ete}
     \end{subfigure}
        \caption{Friedman-Nemenyi test for the anomaly detection methods based on BERT, OpenAI embeddings and end-to-end at significance level 0.1 (the lower the better).}
        % \textcolor{red}{You have only 7 datasets. Compare with 2 to 4 key contenders only.}}
        \label{fig:nem}
\end{figure}

With BERT embeddings, SIK performs best on Email\_Spam, SMS\_Spam, BBC\_News and Movie\_Review datasets. A Friedman-Nemenyi test~\cite{demvsar2006statistical} in Figure~\ref{bert} shows that SIK is top-ranked, only SIK and IDK are significantly better than DeepSVDD, ECOD and Forest, but SIK is much faster than IDK.

SIK shows further performance improvements when applied to OpenAI embeddings, achieving the highest AUROC scores on several datasets, including Email\_Spam, BBC\_News, and N24News. The high-dimensional OpenAI embeddings contain more nuanced semantic information, which SIK successfully leverages for more accurate anomaly detection. Figure~\ref{openai} shows that SIK is top-ranked and the performance of SIK has a critical difference from DeepSVDD and iForest, but IDK doesn't have.

Compared with end-to-end approaches (CVDD, DATE, and FATE), the two-step approach with SIK usually demonstrates superior performance. For instance, on the BBC\_News dataset, OpenAI+SIK significantly outperforms all 3 end-to-end methods. Figure~\ref{ete} shows that SIK is the only detector that significantly better than CVDD.

Compared with other isolation-based approaches (iForest, iNNE and IDK), SIK maintains comparable or superior performance despite its reduced feature dimensionality. With OpenAI embeddings on the SMS\_Spam dataset, SIK achieves higher AUROC than both iForest and IDK, indicating that the SIK preserves the essential discrimination information in the low-dimensional sparse map. 

Since OpenAI-based methods perform better than end-to-end and BERT-based methods, we will focus on OpenAI-based methods in the following subsections.

\subsection{Scalability Analysis}

\textbf{Memory complexity:} SIK achieves space efficiency improvements by focusing solely on boundary information. During training, SIK only needs to store hypersphere information rather than mapping the entire training dataset to feature spaces as required by IDK. During testing, SIK reduces the feature representation dimensionality from $\psi t$ to just $t$, representing a significant reduction in space complexity from $O(nt\psi)$ to $O(nt)$, where the training set has $n$ points.

\textbf{Time complexity:} 
The fundamental difference between SIK and IDK emerges in how they process data during both the training and testing phases. 
During training, SIK directly calculates anomaly scores via norm computations, whereas IDK requires mapping the entire dataset to compute KME, making SIK substantially faster.

During testing, while both SIK and IDK have the same mapping complexity of $O(nt\psi)$, SIK's feature map dimensionality is only $t$ compared to IDK's $\psi t$ dimensions. This dimensional reduction translates to a testing complexity of only $O(nt)$ for SIK versus $O(nt\psi)$ for IDK when computing similarities, resulting in computational savings, particularly for larger values of $\psi$. The overall time complexity is linear because $t \psi$ are hyperparameters and $t \psi \ll n$ for large datasets.

 Table~\ref{tab:scala} presents the runtime and memory costs of both IDK and SIK on the SMS\_Spam dataset with the same hyperparameters $\psi=256, t =200$. SIK completes training approximately 14 times faster than IDK while requiring dramatically less memory. During testing, memory savings remain substantial while time differences are less pronounced.

\begin{table}[!htbp]
\caption{Time and memory comparison on SMS\_Spam where $\psi=256, t=200$. }
% \textcolor{red}{IK $=>$ IDK: so as in main text, e.g., in the section on time complexity}}
\centering
\label{tab:scala}
\resizebox{0.4\textwidth}{!}{%
\begin{tabular}{c|cc|cc}
\hline
      & \multicolumn{2}{c|}{\textbf{Time} (CPU seconds)} & \multicolumn{2}{c}{\textbf{Memory} (MB)} \\
      & IDK           & SIK        & IDK             & SIK       \\ \hline
Train & 115.4       & 8.2       & 1235.2        & 0.5       \\
Test  & 46.3        & 45.6      & 589.8         & 2.3       \\ \hline
\end{tabular}}
\end{table}

Although LOF, ECOD and LUNAR demonstrate good performance on OpenAI embeddings, LOF's quadratic time complexity and LUNAR's computationally intensive deep learning approach result in significantly slower runtime compared to SIK. Table~\ref{runtime} shows the runtime comparison.
% LUNAR, LOF, and  also demonstrate competitive detection performance. SIK achieves extremely fast training because it only samples from the training set and computes pairwise distances among sampled points. While LOF shows comparable testing time to SIK, its training time is significantly longer, which becomes more pronounced on larger datasets. Additionally, SIK supports GPU acceleration, further reducing training time substantially. 

\begin{table}[!htbp]
    \centering
    \caption{Runtime comparison on the Movie Review dataset using default parameters for all methods, and $\psi$=4 and t=100 for SIK in seconds.}
    \resizebox{0.4\textwidth}{!}{
    \begin{tabular}{cccc}
    \toprule
    Methods	 &Train time &	Test time	 &Total time \\
    \midrule
    LUNAR	 &145.1	 &18.0	 &163.1 \\
    LOF	      &20.3	  &10.3	  &30.6 \\
    ECOD	 &18.8	 &28.0	 &46.8  \\
    SIK (CPU)	 &0.1	 &9.5	 &9.6 \\
    SIK (GPU)	 &0.1	 &1.3	 &1.4   \\
    \bottomrule
    \end{tabular}}
    \label{runtime}
\end{table}

\subsection{Sensitivity Analysis}

The performance of SIK depends on two key hyperparameters: the number of hyperspheres per partitioning $\psi$, and the number of partitionings $t$. 
To evaluate this sensitivity, we conduct sensitivity analysis on the \texttt{SMS\_Spam} dataset using OpenAI's \texttt{text-embedding-3-large} embeddings. We vary $\psi \in \{16, 32, 64, 128, 256\}$ with fixed $t=200$, and $t \in \{100, 200, 300, 400, 500\}$ with fixed $\psi=16$. The results are summarized in Table~\ref{tab:sensitivity}.
Our experiments show that SIK is relatively insensitive to $t$ due to the Monte Carlo ensemble effect, averaging over multiple random partitionings stabilizes the anomaly scores. However, SIK is more sensitive to $\psi$, as it directly controls the granularity of the boundary approximation around normal data regions.

\begin{table}[!htbp]
\centering
\caption{Sensitivity analysis on SMS\_Spam.}
\label{tab:sensitivity}
\resizebox{0.45\textwidth}{!}{%
\begin{tabular}{c|ccccc}
\toprule
\textbf{$\psi$} & 16 & 32 & 64 & 128 & 256 \\
\midrule
AUROC & 0.7673 & 0.8054 & 0.8355 & 0.8617 & 0.8967 \\
\midrule
\textbf{$t$} & 100 & 200 & 300 & 400 & 500\\
\midrule
AUROC & 0.7691 & 0.7553 & 0.7577 & 0.7625 & 0.7657 \\
\bottomrule
\end{tabular}}
\end{table}

\subsection{Training with impure data}

% This section examines how robust an anomaly detector is against contamination anomalies in the training set. Figure~\ref{fig:conta} illustrates the performance of SIK on the Email\_Spam dataset with various anomaly ratios. As the proportion of anomalies in the training data increases from 1\% to 5\%, SIK exhibits a remarkably gradual decline in AUROC performance, decreasing from 0.9652 to 0.9528. Despite this slight downward trend, the method consistently maintains excellent detection capabilities, with all performance values remaining above 0.95 AUROC. This demonstrates SIK's robust nature and its capacity to effectively identify anomalies even when the training data contains increasing levels of anomalous contamination.

\begin{figure}[!htbp]
    \centering
    \includegraphics[width=0.9\linewidth]{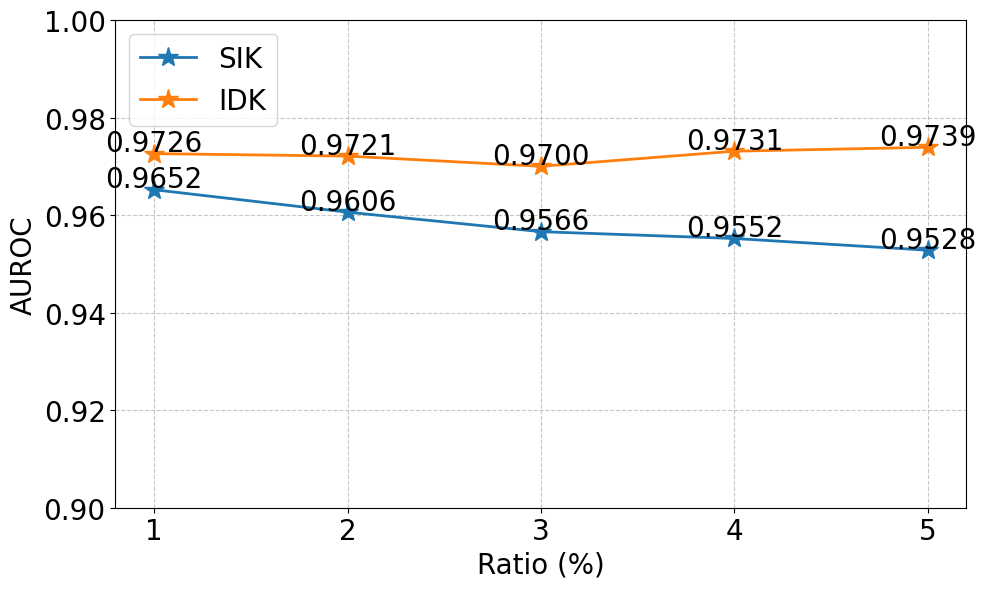}
    \caption{The performance of 5 runs of SIK on Email\_Spam OpenAI embedding. The anomaly ratio is the ratio of the number of normal points and anomalies in the given dataset. }
    % \textcolor{red}{Should include an end-to-end method and a close contender in this evaluation}}
    \label{fig:conta}
\end{figure}

This section examines how robust anomaly detectors are against contamination in training data. Figure~\ref{fig:conta} illustrates the performance of both SIK and IDK on the Email\_Spam dataset with increasing anomaly ratios from 1\% to 5\%. SIK exhibits a gradual decline in AUROC performance, while IDK maintains more stable performance. Despite this slight downward trend, SIK consistently maintains excellent detection capabilities with all values remaining above 0.95 AUROC. 

The performance difference occurs despite both methods using identical hypersphere construction. SIK's decline stems from its reliance on binary inside/outside boundary decisions; when anomalies become hypersphere centers, they create spheres that erroneously encompass other anomalies, misclassifying them as normal. In contrast, IDK demonstrates greater robustness because it goes beyond simple boundary decisions by utilizing kernel mean embedding (KME), which averages feature representations across all training samples. This ensemble effect mitigates the negative impact of individual anomalous sphere centers, allowing IDK's similarity measurement to remain highly effective even when hyperspheres are distorted by contamination. 

\section{Conclusion}

In this paper, we introduced the Simplified Isolation Kernel (SIK) for text anomaly detection. SIK effectively overcomes the computational and memory challenges posed by dense text embeddings from pre-trained LLMs by mapping high-dimensional dense embeddings to a low-dimensional sparse space that preserves boundary information. 
The key innovation of SIK lies in its boundary-focused feature mapping, which maintains a linear time complexity and significantly reduces the dimensionality of the feature representation. 

\newpage

\section*{Limitations}
Although the proposed SIK has shown encouraging results in text anomaly detection, some issues remain for future consideration. While SIK was thoroughly compared with both end-to-end and two-step methods, we did not compare it with direct LLM reasoning approaches due to their significantly slower processing speed and output inconsistencies. Our attempts to use LLMs directly for anomaly detection produced results where the number of output labels frequently mismatched the test data quantity and could not be properly mapped to original text indices, preventing meaningful comparison. 

Additionally, SIK demonstrates effective integration with LLM-generated embeddings, but its applicability to more nuanced domains such as legal, medical, or technical texts requires further investigation. Future work should also explore SIK's capability in detecting subtle anomalies that maintain similar semantic structures to normal text but contain misleading information or factual errors.

\section*{Ethic Statement}

\textbf{Data Sources and Usage:} This study utilizes publicly available research datasets commonly referenced in NLP and anomaly detection literature. All datasets are properly cited throughout the paper. No private, proprietary, or personally identifiable information was included in our research.

\noindent \textbf{Risks and Responsible Use:} While anomaly detection technologies offer valuable capabilities for content moderation and security applications, we recognize they could potentially be misused for surveillance, censorship, or discriminatory filtering. We emphasize that SIK should be deployed responsibly with clear guidelines that respect privacy rights and freedom of expression. The technology presented in this paper is intended for research purposes and legitimate applications such as spam detection, fraud prevention, and identification of harmful content, not for arbitrary surveillance or censorship activities.

\noindent \textbf{Use of AI Assistance:} We acknowledge the use of AI-based writing assistants for grammatical refinement, spelling correction, and improving the clarity of our manuscript. However, all intellectual contributions, experimental designs, analyses, and conclusions in this paper are solely the work of the authors. The development of SIK, its implementation, experimentation, and evaluation were conducted exclusively by the authors without automated generation of scientific content.

\section*{Acknowledgments} This work was supported by National Natural Science Foundation of China (NSFC) (Grant No.62476047), Peking University Mathematics Challenge Funding Program 
(Grant No.2024SRMC10), Guangdong Research Team for Communication and Sensing Integrated with Intelligent Computing (Project No.2024KCXTD047), the Guangdong Provincial Key Laboratory of Mathematical and Neural Dynamical Systems (Grant No.2024B1212010004), the Cross Disciplinary Research Team on Data Science and Intelligent Medicine(Grant No. 2023KCXTD054).

\bibliography{main}

\end{document}